\documentclass[conference]{IEEEtran}
\IEEEoverridecommandlockouts

\usepackage{cite}
\usepackage{amsmath,amssymb,amsfonts}
\usepackage{graphicx}
\usepackage{textcomp}
\usepackage{xcolor}
\usepackage{url}
\usepackage{amsthm}
\usepackage{subfigure}
\usepackage{tabularx}
\usepackage{enumitem}
\usepackage{multicol}
\usepackage{multirow}
\usepackage{algorithm}
\usepackage{hyperref}
\usepackage{booktabs}
\usepackage{soul}
\usepackage{algpseudocode}

\def\BibTeX{{\rm B\kern-.05em{\sc i\kern-.025em b}\kern-.08em
    T\kern-.1667em\lower.7ex\hbox{E}\kern-.125emX}}
\begin{document}

\title{DELTA: Variational Disentangled Learning for Privacy-Preserving Data Reprogramming}


\author{\IEEEauthorblockN{Arun Vignesh Malarkkan}
\IEEEauthorblockA{\textit{Arizona State University}\\
Tempe, Arizona, USA \\
arun.malarkkan@asu.edu}
\and
\IEEEauthorblockN{Haoyue Bai}
\IEEEauthorblockA{\textit{Arizona State University}\\
Tempe, Arizona, USA \\
haoyue.bai@asu.edu}
\and
\IEEEauthorblockN{Anjali Kaushik}
\IEEEauthorblockA{\textit{Arizona State University}\\
Tempe, Arizona, USA \\
akaush39@asu.edu}
\and
\IEEEauthorblockN{Yanjie Fu}
\IEEEauthorblockA{\textit{Arizona State University}\\
Tempe, Arizona, USA \\
yanjie.fu@asu.edu}
}

\maketitle

\begin{abstract}
In real-world applications, domain data often contains identifiable or sensitive attributes, is subject to strict regulations (e.g., HIPAA, GDPR), and requires explicit data feature engineering for interpretability and transparency. 
Existing feature engineering primarily focuses on advancing downstream task performance, often risking privacy leakage. 
We generalize this learning task under such new requirements as Privacy-Preserving Data Reprogramming (PPDR): given a dataset, transforming features to maximize target attribute prediction accuracy while minimizing sensitive attribute prediction accuracy. 
PPDR poses challenges for existing systems: 
1) generating high-utility feature transformations without being overwhelmed by a large search space, and 
2) disentangling and eliminating sensitive information from utility-oriented features to reduce privacy inferability. 
To tackle these challenges, we propose DELTA, a two-phase variational disentangled generative learning framework. 
Phase I uses policy-guided reinforcement learning to discover feature transformations with downstream task utility, without any regard to privacy inferability.
Phase II employs a variational LSTM seq2seq encoder-decoder with a utility-privacy disentangled latent space design and adversarial-causal disentanglement regularization to suppress privacy signals during feature generation. Experiments on eight datasets show DELTA improves predictive performance by $\sim9.3\%$ and reduces privacy leakage by $\sim35\%$, demonstrating robust, privacy-aware data transformation.
\end{abstract}

\section{Introduction}
Data engineering is a cornerstone of domain-inspired Artificial Intelligence (AI), empowering applications across finance, healthcare, and location-based services through the transformation, editing, reprogramming, and augmentation of tailored data representation. 
As these applications increasingly process sensitive information, the need for privacy-preserving data engineering has become indispensable. 
For instance, in the 2018 Strava heatmap leak incident, anonymized fitness data inadvertently exposed secret military base locations. 
This highlights a fundamental risk: optimizing data utility without regard for privacy can lead to unintended data leakage and erode trust in AI systems.
Under this context, we study the Privacy-Preserving Data Reprogramming (PPDR) problem, which aims to reprogram the raw feature representation of a given dataset into a transformed feature set representation that improves target attribute prediction, while safeguarding sensitive attributes. 
Solving PPDR is essential to improve the trustworthiness and deployment of machine learning in sensitive domains. 
For example, in real world domains: 1) raw data often contain personally identifiable or confidential attributes; 2) strict regulatory constraints (e.g., HIPAA~\cite{gostin2009beyond}, GDPR~\cite{zaeem2020effect}) prohibit sharing or storing domain data in original form; 3) instead of relying on anonymization or latent embedding, many domains need explicit interpretable data offering both augmented prediction power and privacy preservation.

There are two major challenges in solving the PPDR problem:
1) Maximize the utility of transformed features, and
2) Minimize the inferability of sensitive attributes from the transformed features.
First, the combinatorial nature of feature transformations in high-dimensional data leads to an exponentially large search space that makes search-based feature engineering methods ineffective and suboptimal.
The first challenge aims to answer: how can we discover high-utility feature transformations without being overwhelmed by the sheer size and complexity of the search space?
Second, transformed features must effectively reduce the sensitive attribute prediction accuracy and minimize privacy leakage. 
The second challenge seeks to answer: how can we detect, identify, and eliminate privacy-related information while preserving utility-oriented feature information during data feature transformation?
Moreover, we need to jointly balance the utility-privacy trade-off within a unified learning framework.

The existing relevant works only partially solve PPDR.
First, PPDR is closely related to automated feature engineering, which includes algorithmic search-based automated feature engineering~\cite{inproceedings, 7836821},  reinforcement learning based automated feature engineering~\cite{liuautodata, ying2025survey, liuRL2019, liuRL2023}, and deep learning based automated feature engineering~\cite{farias2016automatic}.
However, algorithmic search and reinforcement-based methods suffer from a large combinatorial search space;  deep generative learning based methods mainly focus on advancing target attribute prediction accuracy, and lack the ability to explicitly suppress sensitive attribute inference.
Second, PPDR is also related to privacy-preserving embedding or representation learning. 
Prior literature in this area typically entangles utility and privacy objectives into a unified model through adversarial objectives optimization~\cite{edwards2015censoring, madras2018learning, Wu_2022} or injects noise through Differential Privacy~\cite{jiang2022dp2vae}. 
However, these methods generate opaque latent embeddings rather than explicit feature transformations, limiting their interpretability and adoption in regulated domains like healthcare, where practitioners require explicit, traceable, and auditable data features. 
Consequently, the field lacks a method that can both discover high-utility transformations and generate explicit, privacy-preserved features. 
Therefore, we need a novel perspective to effectively formulate and solve high-utility privacy-preserved feature transformation. 


\textbf{Our Insight: An Integrated Variational Generation with Utility-Privacy Disentanglement Perspective.} 
Many applications, such as healthcare and finance, require automated feature engineering to be not only interpretable and utility-driven but also privacy-preserving.
Recent advances~\cite{10.1145/3687485} have regarded feature sets as token sequences. These approaches often use generative models to map these sequences to embedding vectors, subsequently searching and decoding the utility-optimal embedding to obtain transformed feature sets. 
However, the embedding process alone does not guarantee privacy, especially when sensitive attributes remain entangled in the transformation pathways. 
To address the issue, we draw upon two key opportunities from variational autoencoders (VAEs) and disentangled representation learning. 
VAEs enable us to learn a Gaussian-described embedding space and sample the optimal embedding with the highest probability from the encoder distribution, while representation disentanglement allows structured separation of utility-oriented and privacy-sensitive signals.
We introduce two key insights: 
1) sampling the highest probability embedding from the variational seq2seq encoder distribution as optimal feature transformation; 
2) leveraging disentanglement losses during encoding to isolate and remove privacy signals during decoding.

\textbf{Summary of Proposed Solution:} 
Inspired by these findings, we present DELTA, a novel disentangled generative AI framework for privacy-preserving feature transformation. 
DELTA operates in two phases: (1) Policy-Guided Feature Transformation Discovery, and (2) Privacy-aware Generative Data Reprogramming.
In Phase I, we develop an information bottleneck-guided multi-agent RL system to construct a knowledge base of diverse feature transformations as training data. In particular, cascading agents select features and operators to construct new transformations, compute corresponding target attribute prediction accuracy as utility scores, and sensitive attribute prediction accuracy as privacy scores. These transformation paths are encoded into structured token sequences.
In Phase II, we introduce a variational encoder-decoder-evaluator model with utility-privacy disentangled latent spaces for privacy-preserving feature transformation. In particular, the encoder separates utility and privacy-oriented embeddings, while the decoder uses only utility embeddings to generate transformed features. Disentanglement, adversarial, and causal regularization losses ensure that sensitive information is suppressed.
Extensive experiments across eight real-world tabular datasets and twelve competitive baselines demonstrate that DELTA improves downstream F1-score by an average of $\sim9.3\%$ and reduces adversarial sensitive attribute leakage by $\sim35\%$, while incurring only a negligible reduction ($<1.3\%$) from peak utility transformation from phase I.

\noindent\textbf{Contributions:} \emph{Formulation:} We study an important problem: privacy-preserving data reprogramming to augment data features and protect sensitive attributes. \emph{Insights:} We identify two insights: 1)  high-utility feature transformation as high-probability sampling from a learned latent distribution;  2) privacy-preserving feature transformation as embedding disentanglement and privacy leakage mitigation. \emph{Techniques:} We develop an encoder-decoder-evaluator neural learning framework by implementing the first insight via variational LSTM seq-to-seq encoder and the second insight via causal disentangled learning losses. 

\vspace{-0.1cm}
\section{Problem Statement}
Given a dataset with target attributes and sensitive attributes, we aim to transform the original feature set of the dataset into a new feature set, in order to improve downstream task performance while minimizing the inferability of sensitive attributes. 
Formally, let $\mathcal{D} = {(\mathbf{x}_i, y_i, s_i)}_{i=1}^{N}$ with $N$ samples be the dataset, where $\mathbf{x}_i \in \mathbb{R}^K$ is a $K$-dimensional instance vector, $y_i \in \mathcal{Y}$ is a target attribute, and $s_i \in \mathcal{S}$ is a sensitive attribute (e.g., age, race, gender, or location) to be protected from inference. 
The feature transformation process can be described as a function $\mathcal{T}: \mathbb{R}^K \rightarrow \mathbb{R}^d$, such that the optimal transformed features $\mathcal{T}^*(x_i)$ can: 1) maximize predictive utility for $y_i$, i.e., $P(y_i \mid z_i)$ is high; 2) minimize inferability of $s_i$, i.e., $P(s_i \mid z_i)$ is low. This objective is formally given by:
\begin{equation}
\label{eq:privacy_utility_tradeoff}
\resizebox{0.9\linewidth}{!}{ 
$\displaystyle
    \mathcal{T}^* = \arg \max_{\mathcal{T} \in \mathcal{H}} \left( \underbrace{\mathbb{E}_{x,y}\left[\log P(y \mid \mathcal{T}(x))\right]}_{\text{Utility}}
    - \lambda \underbrace{\mathbb{E}_{x,s}\left[\log P(s \mid \mathcal{T}(x))\right]}_{\text{Privacy Leakage}} \right)
$}
\end{equation}
where $\lambda > 0$ is a tunable parameter controlling the privacy-utility trade-off, and $\mathcal{H}$ is the transformation hypothesis space.

\vspace{-0.1cm}
\section{Methodology}

\begin{figure*}[t]
  \centering
  \includegraphics[height=5.8cm, width=\textwidth]{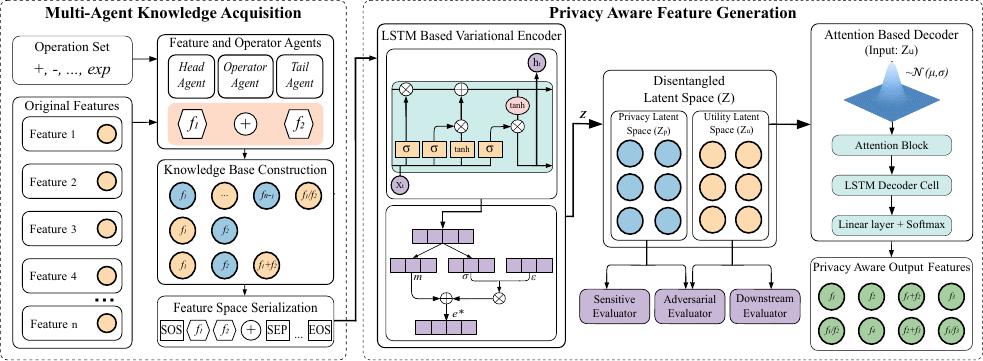}
  \caption{
Overview of the \textbf{DELTA} framework. 
DELTA is a two-phase privacy-preserving feature transformation pipeline. 
Phase I (\textit{Policy-Guided Knowledge Acquisition}) uses reinforcement learning agents, guided by an information bottleneck reward, to explore and acquire utility-driven feature-set transformations as training data for phase II. 
Phase II (\textit{Privacy-Aware Feature Generation}) employs a disentangled variational autoencoder with adversarial and causal regularization to generate transformed features that retain target attribute prediction accuracy while minimizing sensitive feature leakage. 
}
\label{fig:model-architecture}
\vspace{-0.3cm}
\end{figure*}

\noindent Figure \ref{fig:model-architecture} shows our two-phased DELTA framework.

\noindent\textbf{Phase I: Policy-Guided Feature Transformation Discovery.}
We design a multi-agent Reinforcement Learning (RL) system to explore the exponentially large combinatorial space of feature transformations. In particular, the agents are trained to construct feature transformations that maximize predictive utility, guided by an information bottleneck-inspired reward function. 
This yields a knowledge base of transformation sequences annotated with their corresponding utility and privacy leakage scores, which serves as training data for Phase II.

\noindent\textbf{Phase II: Privacy-Enforced Generative Representation Learning.}
Leveraging the curated training data (i.e., feature transformation, utility score, privacy leakage score) from Phase I, we train a variational autoencoder (VAE) with a partitioned latent space to model the generation of new, privacy-preserving feature transformations. This generative model is augmented with adversarial, disentanglement, and causal regularization objectives to disentangle the latent space into a utility-oriented subspace (capturing features that improve target prediction accuracy) and a privacy-oriented subspace (capturing features susceptible to sensitive attribute inference). During generation, we decode feature sets from the utility-oriented embedding of the highest likelihood point to synthesize transformed features, ensuring the output is maximally informative for the prediction task while rigorously limiting sensitive attribute inference.

\vspace{-0.1cm}
\subsection{Phase I: Policy-Guided Feature Transformation Discovery}
The first phase of our framework focuses on collecting training data. Specifically, we seek to discover, for any given dataset, which feature transformations can improve target attribute prediction accuracy under sequential exploration. 
To this end, Phase I develops an automated discovery tool that systematically explores the transformation space and records the utility and privacy impact of each candidate transformation. Inspired by the studies in \cite{dwang_group_fsr,xiao2022traceable}, we formulate reinforcement learning as an automated training data collector and decompose the discovery process into a cascade of interdependent Markov Decision Processes (MDPs), coordinated via a multi-agent Deep Q-Network (DQN) framework. Specifically, 

\subsubsection{Operator and Operand Agents}
We design three MDPs using a cascading structure of three reinforcement learning agents, including two agents (i.e., Head Feature Agent and Tail Feature Agent) to select features and one agent (i.e., Operator Agent) to select operators. Each agent accepts a state and performs actions to select features and mathematical operators for feature crossing, so as to generate new transformed features (e.g. $f_1$/$f_2$, $f_3-f_4$).

\underline{\emph{(1) The Head Feature Agent.}} 
The head feature agent is to select the first feature for feature crossing. At each decision step $t$, the state $s_t$ for the head feature agent is a fixed-length vector encoding the current feature space $\mathcal{F}_t$ and its statistical and structural properties. The action $a_t$ of the head feature agent is to select a single feature from $\mathcal{F}_t$ to serve as the first operand in the feature transformation at step $t$.
Formally, the head feature state $s_t = \psi(\mathcal{F}_t)$, where $\psi$ is \textit{Graph-based Embedding (GCN)} feature embedding function that firstly constructs a feature-feature correlation graph, then embeds it into a state vector via graph convolution.

\underline{\emph{(2) The Operator Agent.}}
The operator agent selects the transformation operator for feature crossing. At each decision step $t$, the state $s_t$ for the operator agent is constructed by concatenating the embedding of the current feature space (via the GCN-based method) with the embedding of the selected head feature. The action $a_t$ is the choice of a transformation operator $o_t \in \mathcal{O}$, where $\mathcal{O}$ denotes the set of available operators (e.g., arithmetic, scaling, nonlinear functions), and each operator is represented by a one-hot encoded vector.

\underline{\emph{(3) The Tail Feature Agent.}}
The tail feature agent selects the second operand for feature crossing. At each step $t$, the agent’s state $s_t$ is formed by concatenating the embeddings of the current feature space and the selected head feature (both via the GCN-based method) with the one-hot embedding of the chosen operator. The action $a_t$ is the selection of a tail feature $f_t$ when the operator is binary; for unary operators, the tail feature agent remains inactive.

\subsubsection{Cooperation between Agents}
We next discuss how to coordinate the three agents for new feature transformation exploration, reward quantification, and DQN training. 

\underline{\emph{(1) New Feature Set Exploration.}}
At the t-th decision step, by applying the operator $o_t$ selected by the operator agent to the two features chosen by the head and tail agents, these agents generate a new feature $f_{new} = o_t(f_h, f_t)$ (or $o_t(f_h)$ for unary). 
The feature set is iteratively updated as $\mathcal{F}_{t+1} = \mathcal{F}_t \cup \{f_{new}\}$. 
By evaluating the predictive power of the new feature set $\mathcal{F}_{t+1}$ on downstream tasks, we calculate a reward score, guiding the agents to explore transformations that maximize predictive performance.

\underline{\emph{(2) Reward and Agent Training.}}
To encourage informative feature sets, we incorporate an information bottleneck principle into the reward function. 
This formulation rewards feature transformations that improve performance on downstream tasks by identifying highly relevant feature representations for the target attribute. 
In particular, we design the reward function to maximize the mutual information between the new feature space and the downstream task labels, which is measured by the incremental improvement in downstream task performance (e.g., F1-score for classification, $1$-RAE for regression) after the transformation.
\begin{equation}
r_t = (1 - \beta_{\text{IB}}) \cdot (\text{Perf}_{t} - \text{Perf}_{t-1})
\end{equation}
where \(\beta_{\text{IB}}\) controls the exploration-utility trade off, \(\text{Perf}_t\) is the performance of the downstream task with the transformed feature set \(\mathcal{F}_t\).
The reinforcement agents aim to maximize the expected cumulative reward, with the policy $\pi$:
\begin{equation}
    J(\pi) = \mathbb{E}_{\pi} \left[ \sum_{t=1}^T r_t \right],
\end{equation}

\underline{\emph{(3) Network Architecture.}}  
Each agent has a personalized DQN to approximate the action-value function. 
For a given agent, the Q-function $Q(s_t, a_t)$ estimates the expected return of taking action $a_t$ in state $s_t$. 
The policy is updated by minimizing the temporal difference (TD) loss:
\begin{equation}
\begin{split}
\mathcal{L}_{\text{TD}} = \mathbb{E}_{(s_t,a_t,r_t,s_{t+1})} \Big[ 
\big( Q(s_t, a_t) \\
- \big( r_t + \gamma \max_{a' \in \mathcal{A}} Q_{\text{target}}(s_{t+1}, a') \big) 
\big)^2 \Big]
\end{split}
\end{equation}
where $\gamma$ is the discount factor and $Q_{\text{target}}$ is a target network periodically synchronized with $Q$. $\epsilon$-greedy exploration is employed to stabilize training and ensure exploration. 

\subsubsection{Collecting Predictive Accuracy of Target and Sensitive Attributes}

To rigorously quantify both utility and privacy, we explicitly evaluate the predictive accuracy of the target attribute and the sensitive attribute:

\underline{\emph{(1) Utility (i.e., target attribute prediction accuracy): }}
For any candidate or generated feature set, we train a downstream classifier (or regressor) to predict the target attribute. The predictive accuracy is measured using standard metrics (e.g., F1-score for classification, $1$-RAE for regression), denoted as $\text{Perf}_t$ in the reward function. This metric directly reflects the utility score $u$ of the feature set for the intended task.

\underline{\emph{(2) Privacy (i.e., sensitive attribute prediction accuracy)}}: To assess privacy leakage, we train a separate classifier to predict the sensitive attribute from the same feature set. The predictive accuracy (e.g., F1-score or accuracy; the lower the better) of this sensitive attribute predictor quantifies the extent to which sensitive information is present in the features (privacy score $p$).

\subsubsection{Feature Space Serialization} 
After exploiting the above reinforcement agents to explore and collect various feature transformations and their corresponding utility and privacy scores, we serialize each transformed feature space as a token sequence $\eta_p$ using a conversion function $\rho(\cdot)$. 
All features and transformation operators are encoded within a unified token vocabulary. 
For each generated feature (e.g., $log((f1+f2)*f3)$), we represent its construction path using Reverse Polish Notation (RPN)~\cite{Krtolica01032004}, to ensure unique and extensible encoding. 
Besides, we leverage special tokens, including $\langle$SOS$\rangle$, $\langle$SEP$\rangle$, and $\langle$EOS$\rangle$, to respectively represent the start, separation, and end of a feature set token sequence (e.g.,$<SOS>f1 f2 + f3 * log<SEP><EOS>)$).


\subsection{Privacy-aware Generative Data Reprogramming}

Phase I develops a method for automated exploration of demonstrations about how different feature transformation paths of a given dataset yield different utility and privacy scores.
Formally, considering the existence of the Phase I-collected training data, where each instance is a tokenized sequence $\mathbf{x} = (x_1, \ldots, x_T)$ that represents a feature transformation trajectory in Reverse Polish Notation (RPN), paired with its corresponding utility $u$ and privacy scores $p$.  
With these demonstrations, Phase II aims to develop a generative feature transformation method to transform the original features of a given dataset to another feature set that maximizes predictive utility while minimizing sensitive attribute leakage by constraining the ability of the newly transformed features to predict sensitive attributes.

\subsubsection{Model Architecture}  
We develop a disentangled sequence-to-sequence variational autoencoder (VAE) framework designed to jointly achieve three core objectives: (1) encoding feature set token sequences to embedding vectors, (2) learning a disentangled latent embedding space that separates utility-oriented and privacy-oriented information, and (3) decoding embedding into the feature set token sequences for generative modeling. Specifically,

\underline{\emph{(1) The Design of Embedding Space:}} 
Rather than mapping feature set token sequences into a single embedding space, we propose to disentangle the latent embedding space into two embedding subspaces: (i) utility-oriented embedding space $z_u$, which encodes target attribute prediction accuracy-related features; (ii) privacy-oriented embedding space $z_p$, which encodes privacy attribute prediction accuracy-related features for privacy control. 
Formally, the entire embedding space $\mathbf{z}$ is a combination of two subspaces, denoted by
$
\mathbf{z} = [\mathbf{z}_u \;\|\, \mathbf{z}_p]$, where $\mathbf{z}_u, \mathbf{z}_p \in \mathbb{R}^{d/2} $, and  $d$ is the total latent dimensionality.

\underline{\emph{(2) The Variational Encoder:}}  
We employ a variational Long Short Term Memory (LSTM) based encoder that embeds a feature set token sequence, where each token denotes an operand or operator in RPN, into an embedding vector. Formally, let $\mathbf{x}$ be a $T$-sized feature set token sequence, its variational LSTM embedding $\mathbf{z} \in \mathbb{R}^{d}$ is given by: 
\begin{align}
\mathbf{e}_{t} &= \operatorname{Embed}(x_{t}), \label{eq:embed}\\
\mathbf{h}_{t} &= \operatorname{LSTM}\!\bigl(\mathbf{e}_{t},\mathbf{h}_{t-1}\bigr), \label{eq:lstm}\\
\mathbf{h}_{\mathrm{pool}} &= \frac{1}{T} \sum_{t=1}^T \mathbf{h}_t, \label{eq:pool}\\
\boldsymbol{\mu} &= W_{\mu}\mathbf{h}_{\text{pool}}+b_{\mu}, \label{eq:mu}\\
\log \boldsymbol{\sigma}^2 &= \mathbf{W}_{\sigma} \mathbf{h}_{\mathrm{pool}} + \mathbf{b}_{\sigma}, \label{eq:logvar}\\
\boldsymbol{\sigma} &= \exp\left(0.5 \log \boldsymbol{\sigma}^2\right), \label{eq:sigma}\\
\boldsymbol{\epsilon} &\sim \mathcal{N}(\mathbf{0}, \mathbf{I}), \label{eq:noise}\\
\mathbf{z} &= \boldsymbol{\mu} + \boldsymbol{\sigma} \odot \boldsymbol{\epsilon}, \label{eq:reparam}
\end{align}
where $\mathbf{e}_t$ is the embedding of the $t$-th token $x_t$, $\mathbf{h}_t$ is the LSTM hidden state,  $\mathbf{h}_{\text{pool}}$ is the mean-pooled embedding of all tokens in a feature set token sequence. $W_{\mu}, W_{\sigma} \in \mathbb{R}^{d \times H}$ are weight matrices, and $b_{\mu}, b_{\sigma} \in \mathbb{R}^{d}$ are bias vectors, in order to parameterise a diagonal Gaussian posterior over the latent space. $\odot$ is elementwise multiplication, $\boldsymbol{\epsilon}$ is standard normal noise.
We then respectively split $\mathbf{z}$ into utility-oriented embedding $\mathbf{z}_u$ and privacy-oriented embedding $\mathbf{z}_p$.

\underline{\emph{(3) Target Attribute Prediction Accuracy Evaluator:}}  
We design a linear layer to take utility-oriented embedding $\mathbf{z}_u$ as input and estimate target attribute prediction accuracy (a.k.a., utility score). Formally, given a feature set token sequence,   its utility-oriented embedding is $\mathbf{z}_u$, its target attribute prediction accuracy $\hat{u}$ is estimated by: 
\begin{align}
\hat{u} = \sigma(\mathbf{w}_{u}^\top \mathbf{z}_u + b_{u}), \label{eq:utility_pred}
\end{align}
where $\mathbf{w}_{u} \in \mathbb{R}^{d/2}$ is the utility-oriented weight vector, $b_{u} \in \mathbb{R}$ is the bias scalar, and $\sigma(\cdot)$ denotes the sigmoid function.

\underline{\emph{(4) Sensitive Attribute Prediction Accuracy Evaluator:}}  
Similarly, we design a linear layer to take the privacy-oriented embedding $z_p$ as input. This component is trained to estimate its privacy value, thereby encouraging $z_p$ to encapsulate sensitive information. Formally, given a feature set token sequence, its privacy-oriented embedding is  $\mathbf{z}_p$, the sensitive attribute prediction accuracy $\hat{p}$ is estimated by:
\begin{align}
\hat{p} = \sigma(\mathbf{w}_{p}^\top \mathbf{z}_p + b_{p}), \label{eq:sensitive_pred}
\end{align}
where $\mathbf{w}_{p} \in \mathbb{R}^{d/2}$ and $b_{p}$ are the privacy-related parameters of a linear function.

\underline{\emph{(5) Adversarial Evaluators:}}  
We introduce two adversarial evaluators for utility-privacy embedding space disentanglement by making the utility-oriented embedding space uninformative on predicting sensitive attributes and making the privacy-oriented embedding uninformative on predicting target attributes.  
In particular,  we design: 1) an \textbf{adversarial sensitive head}, which is to estimate sensitive attribute prediction accuracy using utility-oriented embedding; 2) an \textbf{adversarial utility head}, which is to estimate target attribute prediction accuracy using privacy-oriented embedding. Formally,
\begin{align}
\hat{p}_{\mathrm{adv}} &= \sigma(\mathbf{w}_{\mathrm{adv},p} \mathbf{z}_u + b_{\mathrm{adv},p}), \label{eq:adv_sens}\\
\hat{u}_{\mathrm{adv}} &= \sigma(\mathbf{w}_{\mathrm{adv},u} \mathbf{z}_p + b_{\mathrm{adv},u}), \label{eq:adv_util}
\end{align}
where $\mathbf{w}_{\mathrm{adv},p}, \mathbf{w}_{\mathrm{adv},u} \in \mathbb{R}^{d/2}$ are weight vectors, $b_{\mathrm{adv},p}, b_{\mathrm{adv},u} \in \mathbb{R}$ are bias scalars, and $\sigma(\cdot)$ denotes the sigmoid function.

\underline{\emph{(6) The Decoder:}}  
Our goal is to generate a transformed feature set (a new feature set token sequence) that is uninformative for predicting sensitive attributes. 
During decoding, we thus exclude privacy-oriented embedding and only consider utility-oriented embedding as inputs of an attentive LSTM decoder.
The decoder takes a utility-oriented embedding as its initial state and predicts the next token of maximum probability at each position of the sequence, conditioned on the previous token, the current hidden state, and the attention context vector.
\vspace{-0.1cm}
\begin{align}
\mathbf{h}_0 &= \mathbf{W}_{\mathrm{init}} \mathbf{z}_u + \mathbf{b}_{\mathrm{init}}, \label{eq:dec_init}\\
\mathbf{c}_t &= \operatorname{Attention}(\mathbf{h}_{t-1}, \{\mathbf{h}_1^{\mathrm{enc}}, \ldots, \mathbf{h}_T^{\mathrm{enc}}\}), \label{eq:dec_attn}\\
\left[\mathbf{h}_t, \mathbf{s}_t\right] &= \operatorname{LSTM}\left([\operatorname{Embed}(w_{t-1}); \mathbf{c}_t], \mathbf{h}_{t-1}, \mathbf{s}_{t-1}\right), \label{eq:dec_lstm}\\
\mathbf{\hat{v}}_t &= \operatorname{Softmax}(\mathbf{W}_o \mathbf{h}_t + \mathbf{b}_o), \label{eq:dec_out}
\end{align}
where $\mathbf{z}_u$ is the utility latent vector, $\mathbf{h}_t$ is the decoder hidden state, $\mathbf{c}_t$ is the attention context vector, $w_{t-1}$ is the previously generated token, and $\mathbf{\hat{v}}_t$ is the predicted token probability distribution over the vocabulary at step $t$. 

\subsubsection{Optimization Objective}  
To learn the model structure, we aim to optimize utility, privacy, and disentanglement.

\noindent
\underline{\emph{Incorporating VAE Loss: $\mathcal{L}_{\text{vae}}$.}} We expect the model to learn an effective embedding representation space of feature set token sequences in order to reconstruct the original feature set token sequence and regularize the embedding space as a normal distribution. Therefore, the loss measures two parts of information: i)  how well the decoder can reconstruct the original feature set token sequence from the utility-oriented latent embedding, quantified by a cross-entropy loss between predicted token distributions and  ground-truth sequences, denoted by: $-\frac{1}{N} \sum_{i=1}^N \sum_{t=1}^T \log p_{\theta}(x_{i,t} \mid \mathbf{z}_{u, i})$, where $N$ is the batch size, $T$ is the sequence length, $x_{i,t}$ is the $t$-th token in the $i$-th sequence, $p_{\theta}$ is the decoder’s predicted probability; 
ii) how close the approximate posterior $q(\mathbf{z}|\mathbf{x})$ is close to a standard normal prior $p(\mathbf{z})$, quantified by a KL divergence, denoted by $ -\frac{1}{2} \sum_{j=1}^d \left(1 + \log \sigma_j^2 - \mu_j^2 - \sigma_j^2 \right)$, where  $\mu_j$ and $\sigma_j^2$ are the mean and variance of the $j$-th latent dimension.

\noindent
\underline{\emph{Incorporating Disentanglement Loss: $\mathcal{L}_{\text{disent}}$.}}
We aim to enforce the model to disentangle the utility-oriented embedding space and the sensitive-oriented embedding space, in order to ensure that the utility-oriented embedding is predictive of the target attribute, the privacy-oriented embedding is predictive of the sensitive attribute, and the two embedding spaces are independent. 
In particular, the disentanglement loss aims to: 
i) minimize the error of using utility-oriented embedding to estimate the utility score, measured by binary cross-entropy; 
ii) minimize the error of using privacy-oriented embedding to estimate the privacy score, measured by binary cross-entropy; 
iii) minimize the statistical dependence (covariance) between utility-oriented embedding and privacy-oriented embedding to enforce independence;
iv) maximize the error of adversarial prediction of privacy score using utility-oriented embedding, measured by binary cross-entropy; 
v) maximize the error of adversarial prediction of the utility score using privacy-oriented embedding, measured by binary cross-entropy.

\noindent
\underline{\emph{Incorporating Causal Regularization Loss: $\mathcal{L}_{\text{causal}}$}}
While disentanglement loss enforces statistical independence between utility and sensitive representations, it does not preclude the possibility that changes in the privacy score causally influence the utility-oriented embedding. 
To address this, we introduce a causal regularization loss motivated by~\cite{xiao2023fitness, malarkkan2025rethinking}, that explicitly penalizes any linear influence of privacy score on the utility-oriented latent subspace, thus blocking direct pathways for sensitive information to affect downstream predictions.
Formally, we estimate the optimal linear predictor of the utility embedding $\mathbf{z}_u$ from the sensitive attribute prediction accuracy (i.e., privacy score) $p$ within each batch, and penalize its $\ell_2$ norm:
$
\mathcal{L}_{\text{causal}} = \|\left( (p - \bar{p})^\top (p - \bar{p}) \right)^{-1} (p - \bar{p})^\top (\mathbf{z}_u - \bar{\mathbf{z}}_u)\|_2
$,
where $p$ and $\mathbf{z}_u$ are batch vectors for sensitive attribute prediction accuracy and utility embedding, $\bar{s}$ and $\bar{\mathbf{z}}_u$ their means, and $\|\cdot\|_2$ denotes the $\ell_2$ norm. This loss ensures that the utility representation is robustly protected from direct sensitive attribute effects, making our privacy guarantees resilient even under shifting data distributions.

\vspace{0.1cm}
Finally, the overall objective of our framework is to minimize a weighted sum of the variational encoder-decoder loss, the disentanglement loss, and the causal regularization loss, given by
\begin{align}
\mathcal{L}_{\text{total}} = \mathcal{L}_{\text{vae}} 
+ \lambda_{\text{dis}}\, \mathcal{L}_{\text{dis}} 
+ \lambda_{\text{causal}}\, \mathcal{L}_{\text{causal}}
\end{align}
where $\lambda_{\text{dis}}$, $\lambda_{\text{causal}}$ are hyperparameters.

\vspace{0.3cm}
\subsubsection{Interpretation of Privacy Guarantees}
The objective function delivers robust privacy guarantees through three aspects: \emph{ i) Partitioned Latent Space:}  we separate utility and sensitive information to enable targeted regularization and minimize privacy leakage into utility representations.
 \emph{ ii) Disentanglement and Adversarial Losses:}  
Enforce statistical independence and invariance, ensuring $\mathbf{z}_u$ and $\mathbf{z}_p$ do not encode information about each other's targets, even under adversarial inference.
\emph{ ii) Causal Regularization:}  
By penalizing the regression coefficient $\beta$ in the linear model $\mathbf{z}_u = \beta p + \epsilon$, we eliminate direct linear causal effects of $p$ on $\mathbf{z}_u$ \cite{zhang_causal_fairness}. This provides privacy guarantees robust to distribution shift and adversarial attacks.

\vspace{0.3cm}
\subsubsection{Solving The Optimization Problem}
The model is trained end-to-end using the Adam optimizer. The loss weights for each objective loss:  $\lambda_{\text{dis}}$ and $\lambda_{\text{causal}}$ are selected via cross-validation to balance utility, privacy, and disentanglement objectives. 
To further enhance generalization and robustness, we apply data augmentation techniques such as random shuffling, masking, and disordering to the input feature set token sequences during training.

\vspace{0.3cm}
\subsubsection{Generating Acceptable Optimal Feature Set}
After training the variational encoder, LSTM attentive decoder, and evaluator architecture, the model learns a Gaussian-regularized embedding space, where each embedding represents a possible feature transformation of the input table. 
We propose a three-step approach to generate a transformed feature set to strive for both utility and privacy preservation. In particular, \underline{\emph{Step 1}} is to sample the embedding vector with the highest probability (i.e., Maximum A Posteriori point) in the learned variational Gaussian encoder distribution as the optimal embedding to decode, denoted by $\hat{\mathbf{z}}$. \underline{\emph{Step 2}} is to remove the privacy-oriented embedding $\hat{\mathbf{z}}_p$ from the optimal embedding, and only keep the utility-oriented embedding $\hat{\mathbf{z}}_u$, so that the features decoded from it have a low accuracy in predicting sensitive attributes. In \underline{\emph{Step 3}}, the attentive LSTM decoder conditions on  $\hat{\mathbf{z}}_u$ and the encoder inputs in order to generate the optimal feature transformation token sequence with consideration of both performance and privacy. 
At each decoding step, the model computes a probability distribution over the vocabulary and selects the most probable token via greedy decoding, i.e., taking the $\arg\max$ over the output logits. This process is repeated iteratively to construct the full output sequence until reaching the end of the sequence token.


\vspace{-0.1cm}
\section{Experiments}
We evaluate our method across eight datasets, three types of tasks (i.e., regression, binary classification, and multi-class classification), and five evaluation metrics to answer:
\noindent \textbf{RQ1:} How well does our method perform compared to baselines?
\noindent \textbf{RQ2:} How important is each technical component in our framework?
\noindent \textbf{RQ3:} Can our method trade off between maximizing utility and minimizing sensitive feature leakage?
\noindent \textbf{RQ4:} Does our method achieve disentanglement between utility-oriented embedding and privacy-oriented embedding?
\noindent \textbf{RQ5:} Do our methods generalize across various models?

\vspace{-0.1cm}
\subsection{Experimental Setup}

\begin{table}[th]
\renewcommand{\arraystretch}{1.1}
\centering
\caption{
Dataset Statistics and Sensitive Feature Information. 
}
\vspace{-5pt}
\label{tab:dataset_stats}
\resizebox{\columnwidth}{!}{
\begin{tabular}{l|c|c|c|l}
\hline
\textbf{Dataset} & \textbf{\#Samples} & \textbf{\#Features} & \textbf{Sensitive Feature} & \textbf{Description} \\
\hline
German Credit      & 1001   & 24    & famges      & Marital status \\
Housing Boston     & 506    & 13    & TAX         & Property tax rate \\
UCI Credit Card    & 30000  & 25    & EDUCATION   & Education level \\
Amazon Employee    & 32769  & 9     & ROLE\_CODE  & Role code \\
\hline
Lymphography       & 148    & 18    & --          & -- \\
OpenML 618         & 1000   & 50    & --          & -- \\
AP Omentum Ovary   & 275    & 10936 & --          & -- \\
Activity           & 10299  & 561   & --          & -- \\
\hline
\end{tabular}
}
\end{table}
\subsubsection{Dataset Description}
To cover a broad spectrum of privacy scenarios, dataset sizes, feature dimensionalities, and task types, we select eight diverse domain datasets. 
We first exploit four benchmark datasets (i.e., Housing Boston, German Credit, UCI Credit Card, and Amazon Employee) with identifiable sensitive attributes, such as marital status, credit history, or salary, representing domains where privacy protection is of paramount importance.
To further assess the generalizability of our approach, we then select four additional datasets: Lymphography, OpenML 618, Activity, and AP Omentum Ovary. 
In these cases, a sensitive attribute is randomly selected, simulating scenarios where sensitive features are not explicitly annotated, as in many practical deployments.
Our dataset collection spans a wide range of scales and complexities, from small datasets like Lymphography (148 samples) to large-scale ones such as Amazon Employee (over 32,000 samples), and from low-dimensional (9 features in Amazon Employee) to high-dimensional (over 10,000 features in AP Omentum Ovary) settings. 
The tasks encompass binary and multi-class classification as well as regression, ensuring that our evaluation covers both standard and challenging machine learning problems.
%
Comprehensive statistics and sensitive feature details for all datasets are provided in Tables~\ref{tab:dataset_stats}.


\begin{table*}[t]
\renewcommand{\arraystretch}{1.1}
\centering
\caption{
Performance Evaluation (DT: downstream task F1-Score, higher is better; SF: sensitive feature prediction F1-Score, lower is better). 
\textbf{Bold} indicates the best result if achieved by DELTA, \underline{underline} indicates the best result if achieved by a baseline. 
}
\vspace{-5pt}
\label{tab:all_datasets}
\resizebox{\textwidth}{!}{
\begin{tabular}{l|cc|cc|cc|cc|cc|cc|cc|cc}
\hline
\multirow{2}{*}{\textbf{Method}} 
& \multicolumn{2}{c|}{\textbf{Housing Boston}} 
& \multicolumn{2}{c|}{\textbf{German Credit}} 
& \multicolumn{2}{c|}{\textbf{UCI Credit Card}} 
& \multicolumn{2}{c|}{\textbf{Amazon Employee}}
& \multicolumn{2}{c|}{\textbf{Lymphography}}
& \multicolumn{2}{c|}{\textbf{OpenML 618}}
& \multicolumn{2}{c|}{\textbf{AP Omentum Ovary}}
& \multicolumn{2}{c}{\textbf{Activity}} \\
\cline{2-17}
& DT$\uparrow$ & SF$\downarrow$ 
& DT$\uparrow$ & SF$\downarrow$ 
& DT$\uparrow$ & SF$\downarrow$ 
& DT$\uparrow$ & SF$\downarrow$ 
& DT$\uparrow$ & SF$\downarrow$ 
& DT$\uparrow$ & SF$\downarrow$ 
& DT$\uparrow$ & SF$\downarrow$ 
& DT$\uparrow$ & SF$\downarrow$ \\
\hline \hline
AFAT     & 0.4099 & 0.0359 & 0.7013 & 0.4392 & 0.8056 & 0.9565 & 0.9339 & 0.0381 & 0.6527 & 0.3652 & 0.4741 & 0.0365 & 0.6124 & 0.5743 & 0.9527 & 0.0391 \\
NFS      & 0.4251 & 0.1433 & 0.7061 & 0.4780 & 0.8054 & 0.9531 & 0.9300 & 0.0459 & 0.7180 & 0.5039 & 0.4754 & 0.0420 & 0.6294 & 0.6121 & 0.9506 & 0.0657 \\
TTG      & 0.4140 & 0.1712 & 0.7250 & 0.4499 & 0.7989 & 0.9609 & 0.9316 & 0.0366 & 0.7180 & 0.5501 & 0.4277 & 0.2060 & 0.6345 & 0.6215 & 0.9549 & 0.4361 \\
GRFG     & 0.4212 & 0.1109 & 0.7187 & 0.4555 & 0.8050 & 0.9611 & 0.9309 & 0.0431 & 0.8133 & 0.6323 & 0.4688 & 0.2312 & 0.6443 & 0.6236 & 0.9516 & 0.4559 \\
MOAT     & 0.4648 & 0.0391 & 0.7459 & 0.4432 & 0.8087 & 0.9594 & 0.9344 & 0.0451 & 0.8185 & 0.5100 & 0.4957 & 0.0419 & 0.6713 & 0.6120 & 0.9541 & 0.4515 \\
DP       & 0.4079 & 0.1803 & 0.7080 & 0.4587 & 0.7936 & 0.9682 & 0.9249 & 0.0261 & 0.7175 & 0.6289 & 0.2206 & 0.0363 & 0.6124 & 0.5439 & 0.9488 & 0.4545 \\
GRFG-DP  & 0.4012 & 0.1322 & 0.7005 & 0.4664 & 0.7984 & 0.9670 & 0.9323 & \underline{0.0196} & 0.8138 & 0.2561 & 0.4388 & 0.2329 & 0.6401 & 0.6146 & 0.9518 & 0.4545 \\
MOAT-DP  & 0.4601 & 0.0691 & 0.6905 & 0.4582 & 0.8042 & 0.9637 & 0.9348 & 0.0516 & 0.8180 & 0.4990 & 0.4157 & 0.0369 & 0.6737 & 0.6148 & 0.9522 & 0.4546 \\
RDG      & 0.3761 & 0.1912 & 0.7331 & 0.6113 & 0.6720 & 0.9126 & 0.9287 & 0.0380 & 0.7040 & 0.5111 & 0.4461 & 0.1167 & 0.6622 & 0.4651 & 0.9501 & 0.0401 \\
ERG      & 0.4080 & 0.1103 & 0.7442 & 0.4729 & 0.8030 & 0.8725 & 0.9352 & 0.0239 & 0.6850 & 0.4221 & 0.4621 & 0.0650 & 0.6621 & 0.4450 & 0.9543 & \underline{0.0371} \\
ORI      & 0.4012 & 0.1630 & 0.7012 & 0.4476 & 0.7992 & 0.9665 & 0.9275 & 0.0337 & 0.7175 & 0.4445 & 0.4120 & 0.0423 & 0.6124 & 0.5061 & 0.9503 & 0.0398 \\
\hline
\textbf{DELTA-P1} & \textbf{0.4754} & 0.1524 & \textbf{0.7610} & 0.4120 &\textbf{ 0.8720} & 0.8199 & \textbf{0.9350} & 0.0244 & \textbf{0.8221} & 0.3508 & \textbf{0.5187} & 0.0442 & \textbf{0.6896} & 0.5486 & \textbf{0.9582} & 0.0667 \\
\textbf{DELTA} & 0.4521 & \textbf{0.0313} & 0.7590 &\textbf{ 0.2790} & 0.8246 & \textbf{0.6122} & 0.9326 & \textbf{0.0196} & 0.8199 & \textbf{0.2355} & 0.5182 & \textbf{0.0320} & 0.6713 & \textbf{0.4221} & 0.9521 & 0.0414 \\
\hline
\end{tabular}
}
\end{table*}

\begin{table*}[t]
\renewcommand{\arraystretch}{1.1}
\centering
\caption{
Ablation study of DELTA.
\textbf{w/o CR}: without causal regularization; 
\textbf{w/o Adv}: without adversarial loss; 
\textbf{w/o DL}: without disentanglement loss; 
\textbf{w/o IB}: without information bottleneck reward; 
\textbf{w/o RL Agent}: random or greedy feature selection.
}
\vspace{-5pt}
\label{tab:ablation-study}
\resizebox{\textwidth}{!}{
\begin{tabular}{l|cc|cc|cc|cc|cc|cc|cc|cc}
\hline
\multirow{2}{*}{\textbf{Method}} 
& \multicolumn{2}{c|}{\textbf{Housing Boston}} 
& \multicolumn{2}{c|}{\textbf{German Credit}} 
& \multicolumn{2}{c|}{\textbf{UCI Credit Card}} 
& \multicolumn{2}{c|}{\textbf{Amazon Employee}}
& \multicolumn{2}{c|}{\textbf{Lymphography}}
& \multicolumn{2}{c|}{\textbf{OpenML 618}}
& \multicolumn{2}{c|}{\textbf{AP Omentum Ovary}}
& \multicolumn{2}{c}{\textbf{Activity}} \\
\cline{2-17}
& DT$\uparrow$ & SF$\downarrow$ 
& DT$\uparrow$ & SF$\downarrow$ 
& DT$\uparrow$ & SF$\downarrow$ 
& DT$\uparrow$ & SF$\downarrow$ 
& DT$\uparrow$ & SF$\downarrow$ 
& DT$\uparrow$ & SF$\downarrow$ 
& DT$\uparrow$ & SF$\downarrow$ 
& DT$\uparrow$ & SF$\downarrow$ \\
\hline \hline
DELTA        & 0.4521 & \textbf{0.0313} & 0.7590 &\textbf{ 0.2790} & 0.8246 & \textbf{0.6122} & 0.9261 & \textbf{0.0196} & 0.8199 &\textbf{ 0.2355} & 0.5182 &\textbf{ 0.0320} & 0.6713 & \textbf{0.4221} & 0.9521 & \textbf{0.0414} \\
DELTA-P1     & \textbf{0.4754} & 0.1524 & \textbf{0.7610} & 0.4120 & \textbf{0.8720} & 0.8199 & 0.9350 & 0.0244 & \textbf{0.8221} & 0.3508 & \textbf{0.5187} & 0.0442 & 0.6896 & 0.5486 & \textbf{0.9582} & 0.0667 \\
w/o CR       & 0.4540 & 0.1002 & 0.7350 & 0.3250 & 0.8232 & 0.6799 & 0.9062 & 0.0243 & 0.8201 & 0.3322 & 0.4999 & 0.0442 & 0.6819 & 0.5357 & 0.9555 & 0.0589 \\
w/o Adv      & 0.4711 & 0.1515 & 0.7556 & 0.3789 & 0.8711 & 0.7822 & 0.9299 & 0.0240 & 0.8230 & 0.3323 & 0.5138 & 0.0445 & \textbf{0.6899} & 0.5486 & 0.9580 & 0.0637 \\
w/o DL       & 0.4710 & 0.1555 & 0.7508 & 0.3549 & 0.8616 & 0.6999 & \textbf{0.9351} & 0.0244 & 0.8202 & 0.3323 & 0.5199 & 0.0442 & 0.6883 & 0.5300 & 0.9581 & 0.1640 \\
w/o IB       & 0.4525 & 0.0989 & 0.7120 & 0.3399 & 0.8100 & 0.6989 & 0.9311 & 0.0243 & 0.8221 & 0.3321 & 0.5001 & 0.0441 & 0.6813 & 0.4821 & 0.9221 & 0.1408 \\
w/o RL Agent & 0.4060 & 0.1009 & 0.7212 & 0.2801 & 0.7999 & 0.6701 & 0.9030 & 0.0196 & 0.6641 & 0.2381 & 0.4402 & 0.0354 & 0.6214 & 0.4269 & 0.9529 & 0.0420 \\
\hline
\end{tabular}
}
\end{table*}
\subsubsection{Evaluation Metrics}
We exploit the following metrics: 
\noindent\textbf{1) Utility:} We report \underline{\emph{F1 score}} and \underline{\emph{accuracy}} for classification tasks, and \underline{\emph{ $R^2$}} score for regression tasks, using downstream classifiers (Random Forest, Logistic Regression, SVM, LightGBM, XGBoost) on both original and generated features. Higher F1 scores and accuracy indicate better feature predictive power.
\textbf{2) Privacy Leakage:} We assess the risk of sensitive attribute inference by reporting the \underline{\emph{adversarial accuracy}} and \underline{\emph{AUC-ROC}} of classifiers trained to predict the sensitive attribute from the learned representations or generated features. Lower adversarial accuracy and AUC-ROC indicate better privacy protection
\textbf{3) Disentanglement:} We measure the independence between utility and sensitive representations using \underline{\emph{Hilbert-Schmidt Independence Criterion (HSIC)}} score. Lower HSIC means better disentanglement.

\subsubsection{Baseline Algorithms}
We compare our method with existing feature transformation, selection, and generation algorithms:
\textbf{1) ORG}: Utilizes the unmodified, original dataset for downstream prediction tasks.
\textbf{2) RDG}: Random Data Generation. Constructs feature-operation-data transformation records by randomly selecting operations and features.
\textbf{3) ERG}: Exhaustive Rule Generation. Expands the feature space by systematically applying operations to each feature, followed by a feature selection step to retain the most informative features.
\textbf{4) AFAT}~\cite{horn2019autofeat}: An advanced variant of ERG, incorporating multi-stage feature selection to enhance informativeness.
\textbf{5) NFS}~\cite{chen2019NFS}: Models the transformation process for each feature as a sequence and leverages reinforcement learning to optimize the transformation pipeline.
\textbf{6) TTG}~\cite{khurana2018TTG}: Represents feature transformations as a graph and applies reinforcement learning to efficiently search for optimal transformation paths.
\textbf{7) GRFG}~\cite{dwang_group_fsr}: Employs a collaborative reinforcement learning framework with multiple agents to generate new features.
\textbf{8) MOAT}~\cite{wang2024MOAT}: Employs embedding-optimization-reconstruction paradigm to synthesize quality feature spaces.
Aside from comparison with feature engineering methods, we compare our method with privacy-preserving baselines:
\textbf{9) Data Perturbation (DP)}~\cite{dwork2014DataPer}: Adds calibrated noise to sensitive attributes to obscure them, adhering to differential privacy principles~\cite{dwork2006DP} during both downstream and sensitive attribute prediction tasks.
We also consider hybrid approaches that combine data reprogramming with privacy mechanisms.
\textbf{10) GRFG-DP}: Applies GRFG for feature generation, followed by DP-based perturbation of sensitive features.
\textbf{11) MOAT-DP}: Utilizes MOAT for feature reprogramming, then applies DP to sensitive attributes.
Moreover, we develop a variant of our framework to isolate the contribution of Phase II. \textbf{12) DELTA-P1}:  The utility-only variant that executes \emph{Phase I} (policy-guided RL search) but omits Phase II’s disentanglement and privacy constraints.
For all the methods, we use Random Forests~\cite{breiman2001randomforests} as the downstream predictive model. Its robustness and reliability help to minimize the confounding effects of model variability.

\begin{figure*}
    \centering
    \includegraphics[height=7cm,width=0.9\textwidth]{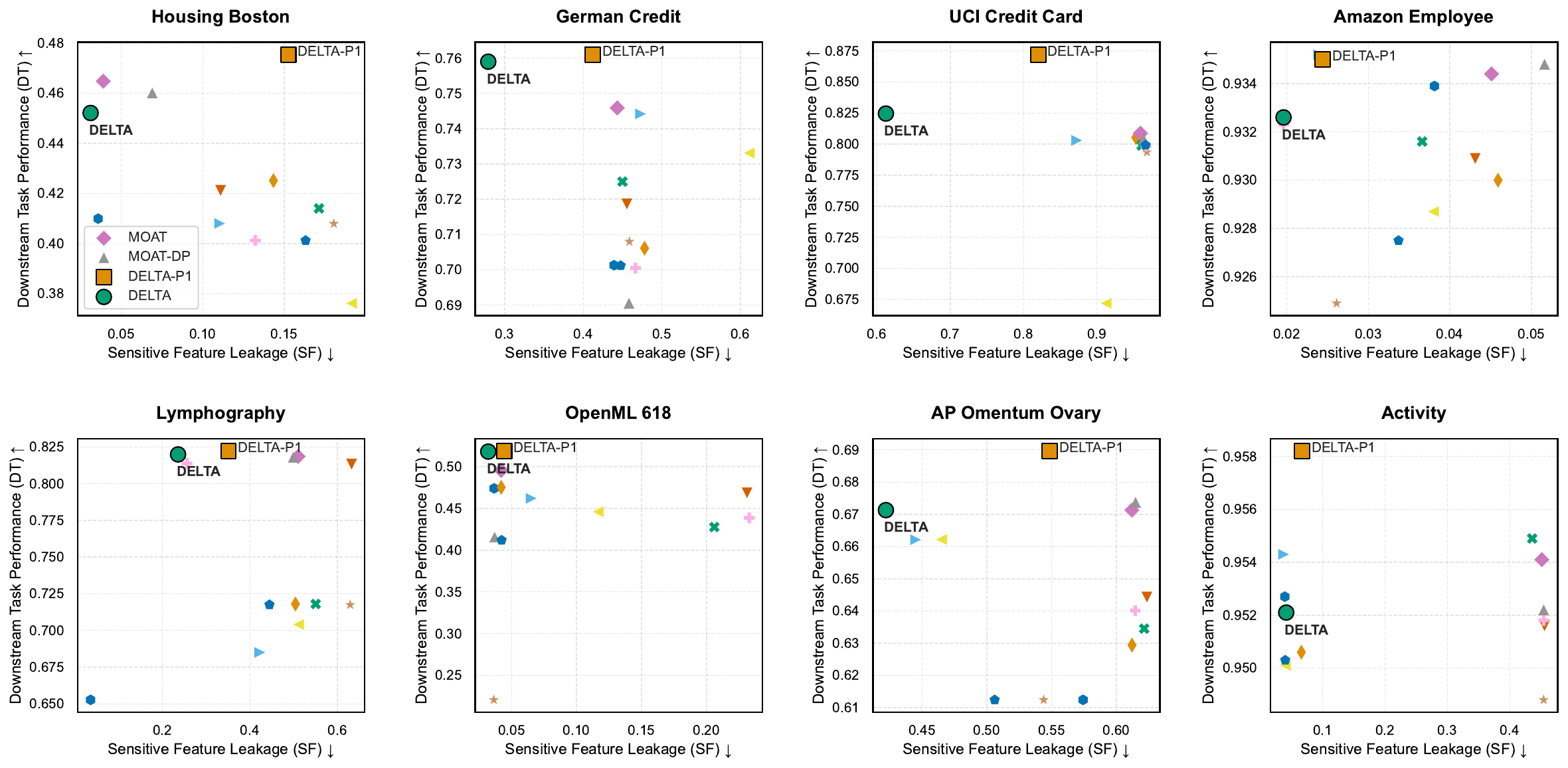}
    \caption{\textbf{Privacy-Utility Trade Off Analysis.}}
    \label{fig:privacy-utility-tradeoff}
\end{figure*}

\vspace{-0.1cm}
\subsection{Experimental Results}

\subsubsection{RQ1: A Study of Overall Performance}
To answer RQ1, we compare DELTA with 11 baseline methods in terms of feature engineering and privacy preservation across 8 benchmark datasets, spanning both classification and regression tasks with a mix of explicit and implicit sensitive attributes. 
Table~\ref{tab:all_datasets} shows the downstream task performance (DT) and sensitive attribute inference (SF), both measured by the F1 score (higher DT and lower SF are desirable). Across all the datasets, \textbf{DELTA achieves an average utility improvement of $\sim9.3\%$ and reduces sensitive attribute leakage by $\sim35\%$ relative to the original dataset features for each task}.
This performance holds across high-dimensional biomedical data (i.e., \textit{AP Omentum Ovary}),  user-centric datasets (i.e., \textit{UCI Credit Card}), and synthetic sensitive attributes (i.e., \textit{OpenML 618}), demonstrating DELTA’s robustness to data modality, dimensionality, and sensitive attribute type. Notably, DELTA consistently outperforms both naive and post-hoc privacy baselines such as GRFG-DP and MOAT-DP, underscoring the advantage of integrated, joint optimization over sequential or disconnected approaches. 
These results validate our central hypothesis: explicitly decoupling utility-driven transformation discovery from privacy-aware generation yields robust, high-utility representations with substantially reduced privacy risk.

\subsubsection{RQ2: Ablation Study}

We present a detailed ablation of DELTA in the table \ref{tab:ablation-study}, highlighting the critical role each component plays in achieving an optimal utility-privacy balance. Removing the privacy-aware generation phase (\textbf{DELTA-P1}) yields the highest utility, but at the cost of substantial sensitive attribute leakage, underscoring the necessity of explicit privacy enforcement. 
Excluding the adversarial (\textbf{w/o Adv}) or disentanglement losses (\textbf{w/o DL}) significantly increases leakage, as these components directly enforce representation invariance to sensitive signals—via prediction suppression and structural separation, respectively. Excluding causal regularization (\textbf{w/o CR}) and the information bottleneck reward (\textbf{w/o IB}) shows moderate degradation in both utility and privacy, with causal regularization proving especially impactful on high-dimensional datasets such as \textit{AP Omentum Ovary}, \textit{Lymphography} by mitigating spurious correlations promoting robust latent space under complex data distributions. We also evaluate the role of the reinforcement learning agent by replacing it with random or greedy feature selection (\textbf{w/o RL Agent}), which leads to a marked drop in both utility and privacy performance. This highlights the importance of DELTA’s policy-guided exploration strategy that jointly optimizes performance and privacy. These findings validate the synergy between adversarial disentanglement, causal regularization, and guided transformation, each indispensable for robust privacy-preserving representation learning.

\subsubsection{RQ3: A Study of Privacy-Utility Trade Off}

Figure~\ref{fig:privacy-utility-tradeoff} presents the privacy-utility tradeoff across all eight datasets, plotting downstream task utility (DT) against sensitive feature leakage (SF) for each method. The optimal region—high utility and low privacy leakage—corresponds to the top-left of each subplot. DELTA consistently dominates this frontier, achieving both superior utility and minimal sensitive feature predictability, reflecting its effectiveness in balancing privacy and performance. In contrast, DELTA-P1, which omits the privacy-aware generation phase, attains higher utility at the cost of substantially increased privacy leakage, underscoring the necessity of explicit privacy constraints. Baselines such as MOAT and GRFG achieve competitive utility but exhibit significantly higher sensitive feature leakage. These findings provide strong empirical evidence that DELTA’s design enables effective privacy preservation while maintaining high task performance across diverse datasets.


\subsubsection{RQ4: A Study of Utility-Privacy Disentanglement}

\begin{figure}
    \centering
    \includegraphics[height=3.5cm, width=\columnwidth]{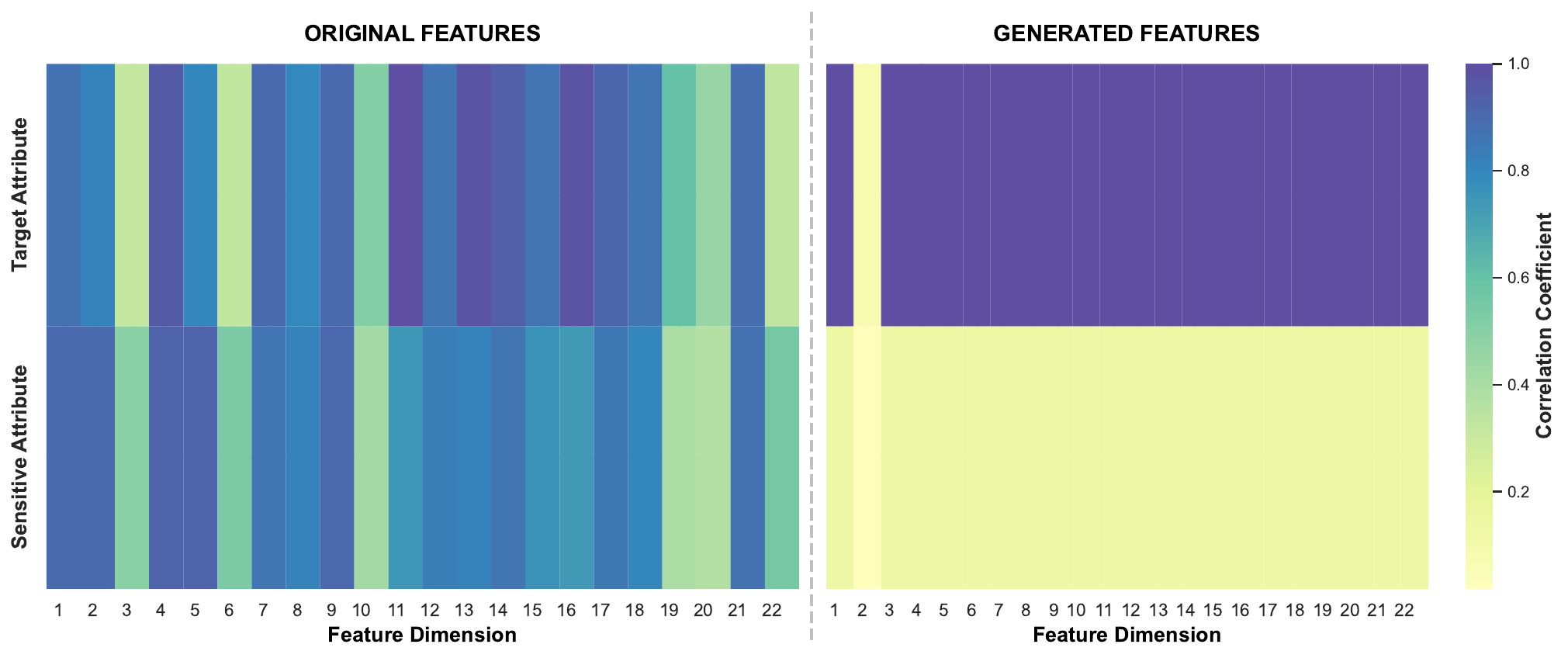}
    \vspace{-0.2cm}
    \caption{\textbf{Utility-Privacy Disentanglement Analysis on German Credit.}}
    \label{fig:dis_hm_analysis}
\end{figure}

To assess the quality of disentanglement in the learned representations, we visualize the heatmaps of feature correlations with both target and sensitive attributes for the original and generated feature transformations on the German Credit dataset. As shown in Figure~\ref{fig:dis_hm_analysis}, the original features exhibit strong correlation with both the target and sensitive attributes, reflecting significant utility-privacy entanglement and elevated risk of privacy leakage. In contrast, DELTA’s generated features exhibit a clear separation: nearly all features are highly correlated with the target attribute while showing minimal correlation with the sensitive attribute, highlighting successful disentanglement. This is further validated with a low HSIC score of 0.0007 between utility and sensitive representations, confirming minimal dependence.

\subsubsection{RQ5: A Study of Cross-Model Generalization}

\begin{figure}
    \centering
    \includegraphics[height=3.5cm, width=\columnwidth]{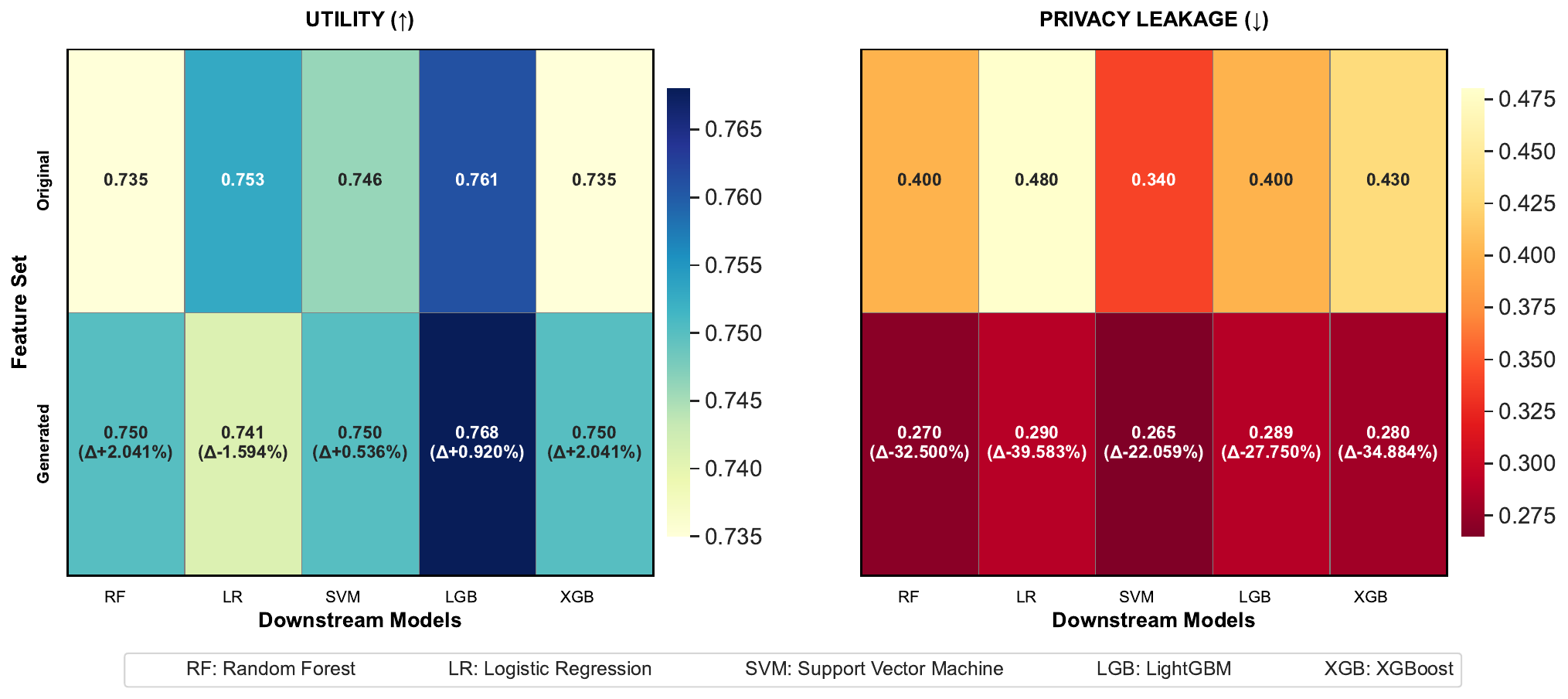}
    \vspace{-0.2cm}
    \caption{
        \textbf{Cross-Model Generalization Analysis of DELTA.}
    }
    \label{fig:cmg_analysis}
\end{figure}

Cross-model generalization results in Fig. \ref{fig:cmg_analysis} reveal that DELTA's generated features consistently outperform original features across all downstream classifiers (RF, LR, SVM, LGB, XGB) on the German Credit dataset, achieving higher utility and significantly lower privacy leakage. Notably, DELTA enhances utility while simultaneously reducing privacy leakage, achieving up to $\sim32\%$ reduction, demonstrating that it effectively balances both objectives without compromise. This model-agnostic effectiveness confirms DELTA's strong transferability and generalization in both utility and privacy dimensions.

\subsubsection{A Study of Scalability}
\begin{figure}
    \centering
    \includegraphics[height=3.5cm, width=\columnwidth]{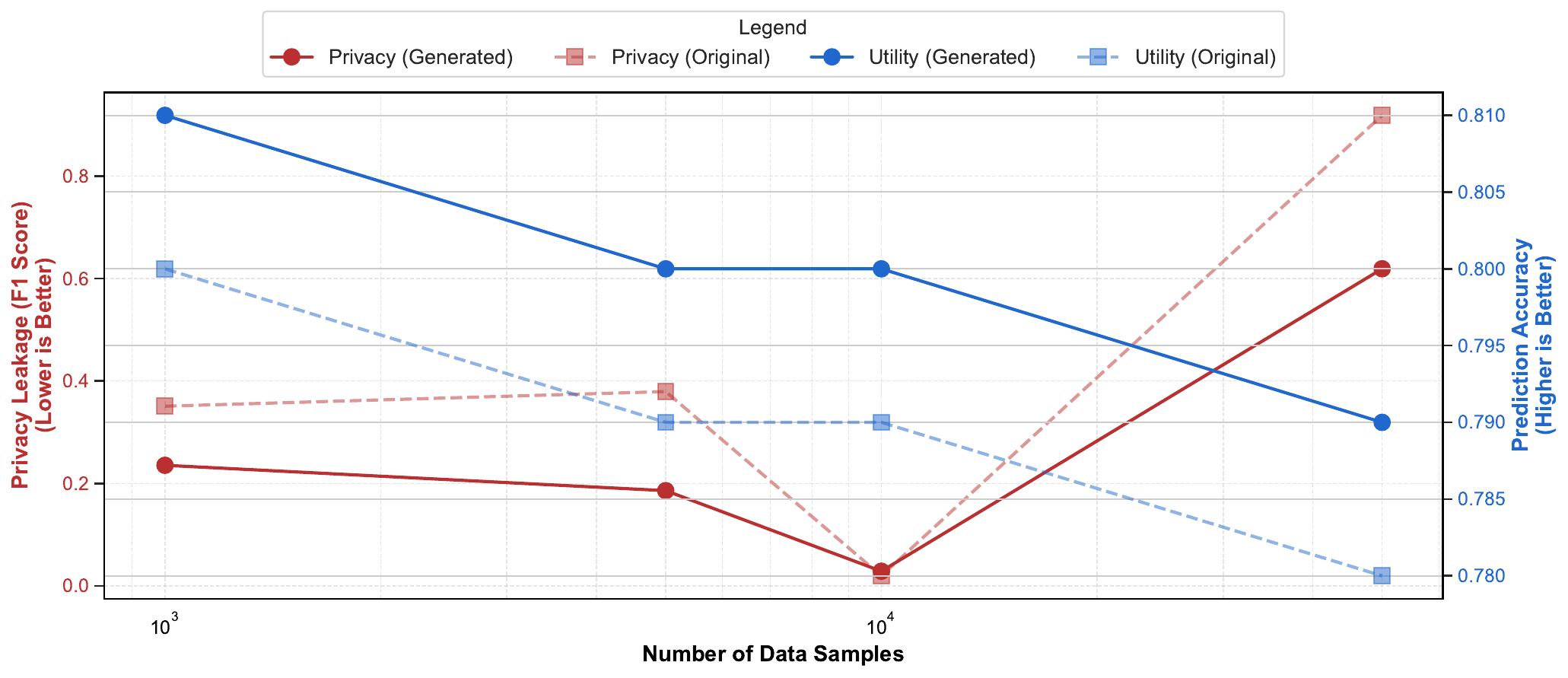}
    \vspace{-0.2cm}
    \caption{
    \textbf{Scalability Performance analysis across different scales.} 
    }
    \label{fig:scalability-DELTA}
\end{figure}

We evaluate DELTA's scalability by measuring runtime and privacy-utility trade-offs on datasets ranging from 1k to 30k samples. Phase 1 runtime increases sublinearly with data size, from 251.5s (1k) to 3629.4s (30k), demonstrating practical efficiency for large-scale applications. Fig. \ref{fig:scalability-DELTA} shows that DELTA-generated features consistently achieve lower privacy leakage than original features, with utility remaining stable across different scales. These results validate that DELTA maintains its privacy-utility advantage and computational tractability as the dataset size grows.

\vspace{-0.2cm}
\section{Related Works}

\noindent{\bf Utility-Optimized Data Reprogramming.}
Data reprogramming is a central task in Data-Centric AI (DCAI) \cite{ying2025survey}, aiming to construct new, traceable features that enhance model performance. Existing approaches typically focus on downstream utility and fall into two main categories:
(1) \underline{\emph{Discrete Space Search}} \cite{dwang_group_fsr,xiao2022traceableautomaticfeaturetransformation, Khurana_Samulowitz_Turaga_2018,7836821,xiao2023traceablegroupwiseselfoptimizingfeature}: These methods perform combinatorial search over transformation sequences, often via greedy iteration \cite{10.1016/j.ins.2011.11.039}, evolutionary strategies \cite{gong2024evolutionarylargelanguagemodel}, or symbolic tracing \cite{xiao2022traceableautomaticfeaturetransformation}. While interpretable and often effective, they suffer from high retraining overhead and limited scalability.
(2) \underline{\emph{Continuous Embedding Optimization}} \cite{wang2023reinforcementenhancedautoregressivefeaturetransformation,10415697,gong2024evolutionarylargelanguagemodel, ying2025distribution}: These methods represent transformation trajectories in a latent space and optimize embeddings to reconstruct high-utility feature sets. Though flexible and expressive, they lack traceability and often ignore privacy constraints.

\noindent{\bf Conditional VAEs for Disentanglement.}
Variational Autoencoders (VAEs) are widely used in DCAI for learning structured, interpretable representations that support fairness, robustness, and debiasing. Disentanglement via VAEs typically follows two paradigms: First, label-driven methods inject supervision into the latent space. A-CVAE \cite{wang2022advancedconditionalvariationalautoencoders} guides disentanglement using class-specific priors but suffers from scalability and generalization issues. Joint label-feature embedding approaches \cite{9898046} assume conditional independence but offer limited separation of confounded factors due to weak disentanglement constraints.
Second, structure-driven models enforce architectural or prior-based regularization without supervision. CVQVAE \cite{ZOU2023172} uses discrete latent codes for controllability, but limits flexibility and diversity. Others like DD-VAE \cite{mathieu2019disentanglingdisentanglementvariationalautoencoders} promote sparse, clustered representations via posterior-prior alignment, but lack principled prior selection and hyperparameter tuning strategies.




\section{Conclusion and Future Work}

This paper advances the frontier of privacy-preserving data-centric AI (DCAI) by addressing the overlooked risks of sensitive attribute leakage in feature transformation.
We proposed \textbf{DELTA}, a novel two-phase framework that decouples utility-driven transformation discovery from privacy-enforced generative modeling. The first phase employs reinforcement learning guided by information bottleneck principles to identify high-utility transformation strategies. The second phase leverages a variational autoencoder with adversarial and disentanglement objectives to generate feature representations that actively mitigate privacy risks while maintaining high utility.
Empirical results show that DELTA consistently enhances predictive performance with significantly reduced privacy leakage. This modular, decoupled design enables systematic exploration of the transformation space under privacy constraints and facilitates causal disentanglement between utility-oriented and sensitive features.
A key limitation is that DELTA’s effectiveness in scenarios with highly imbalanced attribute distributions warrants further investigation.
Future work includes integrating formal guarantees such as differential privacy, scaling to multimodal and streaming data, and adapting DELTA in federated or decentralized learning scenarios. We believe this work lays the groundwork for building safe, accountable, and human-centered AI systems and opens promising directions for privacy-aware data engineering in critical applications.

\bibliographystyle{IEEEtran}
\bibliography{main}

\begin{thebibliography}{10}
\providecommand{\url}[1]{#1}
\csname url@samestyle\endcsname
\providecommand{\newblock}{\relax}
\providecommand{\bibinfo}[2]{#2}
\providecommand{\BIBentrySTDinterwordspacing}{\spaceskip=0pt\relax}
\providecommand{\BIBentryALTinterwordstretchfactor}{4}
\providecommand{\BIBentryALTinterwordspacing}{\spaceskip=\fontdimen2\font plus
\BIBentryALTinterwordstretchfactor\fontdimen3\font minus \fontdimen4\font\relax}
\providecommand{\BIBforeignlanguage}[2]{{%
\expandafter\ifx\csname l@#1\endcsname\relax
\typeout{** WARNING: IEEEtran.bst: No hyphenation pattern has been}%
\typeout{** loaded for the language `#1'. Using the pattern for}%
\typeout{** the default language instead.}%
\else
\language=\csname l@#1\endcsname
\fi
#2}}
\providecommand{\BIBdecl}{\relax}
\BIBdecl

\bibitem{gostin2009beyond}
L.~O. Gostin, L.~A. Levit, and S.~J. Nass, ``Beyond the hipaa privacy rule: enhancing privacy, improving health through research,'' 2009.

\bibitem{zaeem2020effect}
R.~N. Zaeem and K.~S. Barber, ``The effect of the gdpr on privacy policies: Recent progress and future promise,'' \emph{ACM Transactions on Management Information Systems (TMIS)}, vol.~12, no.~1, pp. 1--20, 2020.

\bibitem{inproceedings}
M.~Smith and L.~Bull, ``Feature construction and selection using genetic programming and a genetic algorithm,'' 04 2003.

\bibitem{7836821}
U.~Khurana, D.~Turaga, H.~Samulowitz, and S.~Parthasrathy, ``Cognito: Automated feature engineering for supervised learning,'' in \emph{2016 IEEE 16th International Conference on Data Mining Workshops (ICDMW)}, 2016, pp. 1304--1307.

\bibitem{liuautodata}
J.~Liu, F.~Zhu, C.~Chai, Y.~Luo, and N.~Tang, ``Automatic data acquisition for deep learning,'' \emph{Proc. VLDB Endow.}, vol.~14, no.~12, p. 2739–2742, Jul. 2021.

\bibitem{ying2025survey}
W.~Ying, C.~Wei, N.~Gong, X.~Wang, H.~Bai, A.~V. Malarkkan, S.~Dong, D.~Wang, D.~Zhang, and Y.~Fu, ``A survey on data-centric ai: Tabular learning from reinforcement learning and generative ai perspective,'' \emph{arXiv preprint arXiv:2502.08828}, 2025.

\bibitem{liuRL2019}
K.~Liu, Y.~Fu, P.~Wang, L.~Wu, R.~Bo, and X.~Li, ``Automating feature subspace exploration via multi-agent reinforcement learning,'' in \emph{Proceedings of the 25th ACM SIGKDD International Conference on Knowledge Discovery \& Data Mining}, ser. KDD '19.\hskip 1em plus 0.5em minus 0.4em\relax New York, NY, USA: Association for Computing Machinery, 2019, p. 207–215.

\bibitem{liuRL2023}
K.~Liu, Y.~Fu, L.~Wu, X.~Li, C.~Aggarwal, and H.~Xiong, ``Automated feature selection: A reinforcement learning perspective,'' \emph{IEEE Transactions on Knowledge and Data Engineering}, vol.~35, no.~3, pp. 2272--2284, 2023.

\bibitem{farias2016automatic}
G.~Farias, S.~Dormido-Canto, J.~Vega, G.~Ratt{\'a}, H.~Vargas, G.~Hermosilla, L.~Alfaro, and A.~Valencia, ``Automatic feature extraction in large fusion databases by using deep learning approach,'' \emph{Fusion Engineering and Design}, vol. 112, pp. 979--983, 2016.

\bibitem{edwards2015censoring}
H.~Edwards and A.~Storkey, ``Censoring representations with an adversary,'' \emph{arXiv preprint arXiv:1511.05897}, 2015.

\bibitem{madras2018learning}
D.~Madras, E.~Creager, T.~Pitassi, and R.~Zemel, ``Learning adversarially fair and transferable representations,'' in \emph{International Conference on Machine Learning}.\hskip 1em plus 0.5em minus 0.4em\relax PMLR, 2018, pp. 3384--3393.

\bibitem{Wu_2022}
Q.~Wu, J.~Tang, S.~Dang, and G.~Chen, ``Data privacy and utility trade-off based on mutual information neural estimator,'' \emph{Expert Systems with Applications}, vol. 207, p. 118012, Nov. 2022.

\bibitem{jiang2022dp2vae}
D.~Jiang, G.~Zhang, M.~Karami, X.~Chen, Y.~Shao, and Y.~Yu, ``Dp$^2$-vae: Differentially private pre-trained variational autoencoders,'' 2022.

\bibitem{10.1145/3687485}
W.~Ying, D.~Wang, H.~Chen, and Y.~Fu, ``Feature selection as deep sequential generative learning,'' \emph{ACM Trans. Knowl. Discov. Data}, vol.~18, no.~9, Oct. 2024.

\bibitem{dwang_group_fsr}
D.~Wang, Y.~Fu, K.~Liu, X.~Li, and Y.~Solihin, ``Group-wise reinforcement feature generation for optimal and explainable representation space reconstruction,'' 2022.

\bibitem{xiao2022traceable}
M.~Xiao, D.~Wang, M.~Wu, Z.~Qiao, P.~Wang, K.~Liu, Y.~Zhou, and Y.~Fu, ``Traceable automatic feature transformation via cascading actor-critic agents,'' 2022.

\bibitem{Krtolica01032004}
P.~V. Krtolica and P.~S.~S. and, ``Reverse polish notation method,'' \emph{International Journal of Computer Mathematics}, vol.~81, no.~3, pp. 273--284, 2004.

\bibitem{xiao2023fitness}
Y.~Xiao, S.~Wang, S.~Liu, D.~Xue, X.~Zhan, and Y.~Liu, ``Fitness: A causal de-correlation approach for mitigating bias in machine learning software,'' 2023.

\bibitem{malarkkan2025rethinking}
A.~V. Malarkkan, H.~Bai, X.~Wang, A.~Kaushik, D.~Wang, and Y.~Fu, ``Rethinking spatio-temporal anomaly detection: A vision for causality-driven cybersecurity,'' \emph{arXiv preprint arXiv:2507.08177}, 2025.

\bibitem{zhang_causal_fairness}
J.~Zhang and E.~Bareinboim, ``Fairness in decision-making — the causal explanation formula,'' in \emph{Proceedings of the Thirty-Second AAAI Conference on Artificial Intelligence and Thirtieth Innovative Applications of Artificial Intelligence Conference and Eighth AAAI Symposium on Educational Advances in Artificial Intelligence}, ser. AAAI'18/IAAI'18/EAAI'18.\hskip 1em plus 0.5em minus 0.4em\relax AAAI Press, 2018.

\bibitem{horn2019autofeat}
\BIBentryALTinterwordspacing
F.~Horn, R.~Pack, and M.~Rieger, \emph{The autofeat Python Library for Automated Feature Engineering and Selection}.\hskip 1em plus 0.5em minus 0.4em\relax Springer International Publishing, 2020, p. 111–120. [Online]. Available: \url{http://dx.doi.org/10.1007/978-3-030-43823-4_10}
\BIBentrySTDinterwordspacing

\bibitem{chen2019NFS}
\BIBentryALTinterwordspacing
X.~Chen, B.~Qiao, W.~Zhang, W.~Wu, M.~Chintalapati, D.~Zhang, Q.~Lin, C.~Luo, X.~Li, H.~Zhang, Y.~Xu, Y.~Dang, K.~Sui, and X.~Zhang, ``Neural feature search: {A} neural architecture for automated feature engineering,'' in \emph{2019 {IEEE} International Conference on Data Mining, {ICDM} 2019, Beijing, China, November 8-11, 2019}, J.~Wang, K.~Shim, and X.~Wu, Eds.\hskip 1em plus 0.5em minus 0.4em\relax {IEEE}, 2019, pp. 71--80. [Online]. Available: \url{https://doi.org/10.1109/ICDM.2019.00017}
\BIBentrySTDinterwordspacing

\bibitem{khurana2018TTG}
\BIBentryALTinterwordspacing
U.~Khurana, H.~Samulowitz, and D.~S. Turaga, ``Feature engineering for predictive modeling using reinforcement learning,'' in \emph{Proceedings of the Thirty-Second {AAAI} Conference on Artificial Intelligence, (AAAI-18), the 30th innovative Applications of Artificial Intelligence (IAAI-18), and the 8th {AAAI} Symposium on Educational Advances in Artificial Intelligence (EAAI-18), New Orleans, Louisiana, USA, February 2-7, 2018}, S.~A. McIlraith and K.~Q. Weinberger, Eds.\hskip 1em plus 0.5em minus 0.4em\relax {AAAI} Press, 2018, pp. 3407--3414. [Online]. Available: \url{https://doi.org/10.1609/aaai.v32i1.11678}
\BIBentrySTDinterwordspacing

\bibitem{wang2024MOAT}
D.~Wang, M.~Xiao, M.~Wu, Y.~Zhou, Y.~Fu \emph{et~al.}, ``Reinforcement-enhanced autoregressive feature transformation: Gradient-steered search in continuous space for postfix expressions,'' \emph{Advances in Neural Information Processing Systems}, vol.~36, 2024.

\bibitem{dwork2014DataPer}
\BIBentryALTinterwordspacing
C.~Dwork, K.~Talwar, A.~Thakurta, and L.~Zhang, ``Analyze gauss: optimal bounds for privacy-preserving principal component analysis,'' in \emph{Symposium on Theory of Computing, {STOC} 2014, New York, NY, USA, May 31 - June 03, 2014}, D.~B. Shmoys, Ed.\hskip 1em plus 0.5em minus 0.4em\relax {ACM}, 2014, pp. 11--20. [Online]. Available: \url{https://doi.org/10.1145/2591796.2591883}
\BIBentrySTDinterwordspacing

\bibitem{dwork2006DP}
\BIBentryALTinterwordspacing
C.~Dwork, ``Differential privacy,'' in \emph{Automata, Languages and Programming, 33rd International Colloquium, {ICALP} 2006, Venice, Italy, July 10-14, 2006, Proceedings, Part {II}}, ser. Lecture Notes in Computer Science, M.~Bugliesi, B.~Preneel, V.~Sassone, and I.~Wegener, Eds., vol. 4052.\hskip 1em plus 0.5em minus 0.4em\relax Springer, 2006, pp. 1--12. [Online]. Available: \url{https://doi.org/10.1007/11787006\_1}
\BIBentrySTDinterwordspacing

\bibitem{breiman2001randomforests}
L.~Breiman, ``Random forests,'' \emph{Machine learning}, vol.~45, pp. 5--32, 2001.

\bibitem{xiao2022traceableautomaticfeaturetransformation}
\BIBentryALTinterwordspacing
M.~Xiao, D.~Wang, M.~Wu, Z.~Qiao, P.~Wang, K.~Liu, Y.~Zhou, and Y.~Fu, ``Traceable automatic feature transformation via cascading actor-critic agents,'' 2022. [Online]. Available: \url{https://arxiv.org/abs/2212.13402}
\BIBentrySTDinterwordspacing

\bibitem{Khurana_Samulowitz_Turaga_2018}
\BIBentryALTinterwordspacing
U.~Khurana, H.~Samulowitz, and D.~Turaga, ``Feature engineering for predictive modeling using reinforcement learning,'' \emph{Proceedings of the AAAI Conference on Artificial Intelligence}, vol.~32, no.~1, Apr. 2018. [Online]. Available: \url{https://ojs.aaai.org/index.php/AAAI/article/view/11678}
\BIBentrySTDinterwordspacing

\bibitem{xiao2023traceablegroupwiseselfoptimizingfeature}
\BIBentryALTinterwordspacing
M.~Xiao, D.~Wang, M.~Wu, K.~Liu, H.~Xiong, Y.~Zhou, and Y.~Fu, ``Traceable group-wise self-optimizing feature transformation learning: A dual optimization perspective,'' 2023. [Online]. Available: \url{https://arxiv.org/abs/2306.16893}
\BIBentrySTDinterwordspacing

\bibitem{10.1016/j.ins.2011.11.039}
\BIBentryALTinterwordspacing
O.~Dor and Y.~Reich, ``Strengthening learning algorithms by feature discovery,'' \emph{Inf. Sci.}, vol. 189, p. 176–190, Apr. 2012. [Online]. Available: \url{https://doi.org/10.1016/j.ins.2011.11.039}
\BIBentrySTDinterwordspacing

\bibitem{gong2024evolutionarylargelanguagemodel}
\BIBentryALTinterwordspacing
N.~Gong, C.~K. Reddy, W.~Ying, H.~Chen, and Y.~Fu, ``Evolutionary large language model for automated feature transformation,'' 2024. [Online]. Available: \url{https://arxiv.org/abs/2405.16203}
\BIBentrySTDinterwordspacing

\bibitem{wang2023reinforcementenhancedautoregressivefeaturetransformation}
\BIBentryALTinterwordspacing
D.~Wang, M.~Xiao, M.~Wu, P.~Wang, Y.~Zhou, and Y.~Fu, ``Reinforcement-enhanced autoregressive feature transformation: Gradient-steered search in continuous space for postfix expressions,'' 2023. [Online]. Available: \url{https://arxiv.org/abs/2309.13618}
\BIBentrySTDinterwordspacing

\bibitem{10415697}
W.~Ying, D.~Wang, K.~Liu, L.~Sun, and Y.~Fu, ``Self-optimizing feature generation via categorical hashing representation and hierarchical reinforcement crossing,'' in \emph{2023 IEEE International Conference on Data Mining (ICDM)}, 2023, pp. 748--757.

\bibitem{ying2025distribution}
W.~Ying, N.~Gong, D.~Wang, X.~Wang, A.~V. Malarkkan, V.~Gupta, C.~K. Reddy, and Y.~Fu, ``Distribution shift aware neural tabular learning,'' \emph{arXiv preprint arXiv:2508.19486}, 2025.

\bibitem{wang2022advancedconditionalvariationalautoencoders}
\BIBentryALTinterwordspacing
Y.~Wang, J.~Liao, H.~Yu, G.~Wang, X.~Zhang, and L.~Liu, ``Advanced conditional variational autoencoders (a-cvae): Towards interpreting open-domain conversation generation via disentangling latent feature representation,'' 2022. [Online]. Available: \url{https://arxiv.org/abs/2207.12696}
\BIBentrySTDinterwordspacing

\bibitem{9898046}
K.~Zou, S.~Faisan, F.~Heitz, and S.~Valette, ``Joint disentanglement of labels and their features with vae,'' in \emph{2022 IEEE International Conference on Image Processing (ICIP)}, 2022, pp. 1341--1345.

\bibitem{ZOU2023172}
------, ``Disentangling high-level factors and their features with conditional vector quantized vaes,'' \emph{Pattern Recognition Letters}, vol. 172, pp. 172--180, 2023.

\bibitem{mathieu2019disentanglingdisentanglementvariationalautoencoders}
\BIBentryALTinterwordspacing
E.~Mathieu, T.~Rainforth, N.~Siddharth, and Y.~W. Teh, ``Disentangling disentanglement in variational autoencoders,'' 2019. [Online]. Available: \url{https://arxiv.org/abs/1812.02833}
\BIBentrySTDinterwordspacing

\end{thebibliography}

\vfill

\end{document}